\pdfoutput=1

\documentclass[11pt]{article}

\usepackage{acl}

\usepackage{times}
\usepackage{latexsym}
\usepackage{lipsum}  
\usepackage{graphicx}
\usepackage{subcaption}
\usepackage[T1]{fontenc}

\usepackage[utf8]{inputenc}

\usepackage{microtype}

\usepackage{inconsolata}
\usepackage{array,multirow}
\usepackage{hyperref}
\usepackage{soul}
\usepackage{color}
\usepackage{todonotes}

\title{Large Language Models are Few-Shot Training Example Generators:\\ A Case Study in Fallacy Recognition}

\author{Tariq Alhindi$^1$ \hspace{0.5cm}
        Smaranda Muresan$^2$ \hspace{0.5cm}
        Preslav Nakov$^1$\\
  $^1$Mohamed bin Zayed University of Artificial Intelligence, UAE\\
  $^2$Columbia University, USA\\
  {\normalsize \tt \{tariq.alhindi, preslav.nakov\}@mbzuai.ac.ae} \\
  {\normalsize \tt smara@columbia.edu}}

\begin{document}
\maketitle

\begin{abstract}
Recognizing fallacies is crucial for ensuring the quality and validity of arguments across various domains. However, computational fallacy recognition faces challenges due to the diverse genres, domains, and types of fallacies found in datasets. This leads to a highly multi-class, and even multi-label, setup with substantial class imbalance. In this study, we aim to enhance existing models for fallacy recognition by incorporating additional context and by leveraging large language models to generate synthetic data, thus increasing the representation of the infrequent classes. We experiment with GPT-3.5 to generate synthetic examples and we examine the impact of prompt settings for this. Moreover, we explore zero-shot and few-shot scenarios to evaluate the effectiveness of using the generated examples for training smaller models within a unified fallacy recognition framework. Furthermore, we analyze the overlap between the synthetic data and existing fallacy datasets. Finally, we investigate the usefulness of providing supplementary context for detecting fallacy types that need such context, e.g., diversion fallacies. Our evaluation results demonstrate consistent improvements across fallacy types, datasets, and generators. The code and the synthetic datasets are all publicly available\footnote{\url{https://github.com/Tariq60/fallacy-detection}}.
\end{abstract}

\section{Introduction}
Fallacies are common errors in reasoning that can mislead and invalidate arguments. The capacity to discern fallacies is fundamental to sustaining the robustness and authenticity of arguments across various domains, such as public policy, legal reasoning, and scientific discourse \cite{bailin2016reason}. In recent years, the task of automated fallacy recognition has attracted significant interest from researchers in the fields of Natural Language Processing (NLP) and Artificial Intelligence (AI) \cite{amgoud2013logical,hamblin2022fallacies,goffredo2022fallacious,alhindi-etal-2022-multitask,jin-etal-2022-logical}. However, numerous challenges persist, including the multiplicity of genres, domains, and fallacy types, which contribute to a complex multi-class and multi-label task structure compounded by class imbalances in datasets.

Existing work on fallacy recognition is still in its early stages, with limited datasets available. These datasets cover different types of fallacies in various contexts, such as question and answer dialog moves \cite{habernal-etal-2017-argotario}, name-calling in social media debates \cite{habernal-etal-2018-name}, logical fallacies from educational websites \cite{jin-etal-2022-logical}, and fallacies related to Covid-19 misinformation in social media and news \cite{musi2022developing}. 
Previous work has focused on detecting fallacies in individual datasets, using techniques like fine-tuning transformers for sequence tagging \cite{goffredo2022fallacious}, and training structure-aware classifiers \cite{jin-etal-2022-logical}. However, fallacy recognition is challenging due to the high number of classification labels, class imbalance in datasets, limited dataset sizes, and poor out-of-distribution generalization. 

\citet{alhindi-etal-2022-multitask} proposes a multitask framework using T5, which converts fallacy types into natural language instructions, and thus approaches the differences between fallacy datasets as different tasks, but their approach does not detect infrequent classes effectively. \citet{goffredo2022fallacious} incorporate argumentation features to detect fallacies in political debates, while \citet{jin-etal-2022-logical} trains a structure-aware classifier on fallacies from educational websites; however, they both focus on a single fallacy scheme from one dataset while we include multiple fallacy schemes. 

\begin{figure*}[th!]
    \centering
    \includegraphics[width=\linewidth]{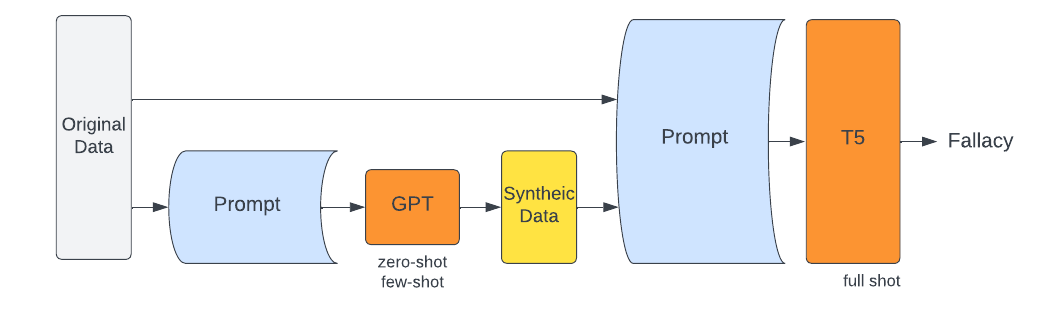}
    \caption{Data augmentation and model training pipeline.}
    \label{fig:intro}
\end{figure*} 

Current state-of-the-art models struggle with the recognition of underrepresented fallacies, which may often require additional context for accurate identification, such as diversion fallacies \cite{walton1996argument}. 

This necessitates a comprehensive and diverse dataset for training these models. One strategy to combat the challenge of sparse and imbalanced data in machine learning is data augmentation \cite{wang2017effectiveness} by creating synthetic examples, thereby enhancing the dataset size and diversity and improving the performance of the machine learning model.

Large Language Models (LLMs) such as GPT-3, 3.5, 4 \cite{brown2020language,achiam2023gpt}, Llama \cite{touvron2023llama} and Mistral \cite{jiang2023mistral} have shown promising zero-shot performance on various text classification tasks \cite{bubeck2023sparks,gilardi2023chatgpt}. However, they still do not perform as good as smaller models that are fine-tuned for specific tasks, especially ones that require deeper understanding and reasoning \cite{alhindi-etal-2022-multitask,qin2023chatgpt}. For complex tasks, such as fallacy recognition, they can be used for data augmentation by bootstrapping existing human-annotated datasets by annotating or generating additional examples \cite{moller2023prompt,lin2023selective,zhu2023can}.


We extend previous work on 
fallacy recognition by exploring the capabilities of large language models to generate synthetic data that augments manually labeled datasets (Section \ref{sec:gen}). We study the effect of the data generated under zero-shot and few-shot conditions on the downstream task of fallacy recognition and show our experimental setup (Section \ref{sec:exp}) and results (Section \ref{sec:res}). We also analyze the quality of the synthetic data and its similarity to the fallacy datasets (Section \ref{sec:analysis}). Figure \ref{fig:intro} shows an overview of our approach of using GPT-3.5 to generate additional training examples in zero/few-shot settings, then training a T5 model \cite{raffel2020exploring} for fallacy recognition on a combination of the original and the synthetic data.


\section{Fallacy Datasets}
\label{sec:dataset}
We experiment with the five fallacy datasets covered by \citet{alhindi-etal-2022-multitask}. They include fallacies in question-answer pairs in game settings (\textsc{Argotario}) \cite{habernal-etal-2017-argotario}, 18 propaganda techniques in news articles (\textsc{Propaganda}) \cite{da-san-martino-etal-2019-fine}, logical fallacies from educational websites (\textsc{Logic}) \cite{jin-etal-2022-logical}, and fallacies in misinformation around covid in social media (\textsc{Covid}) \cite{musi2022developing} and climate change news articles (\textsc{Climate}) \cite{alhindi-etal-2022-multitask}. 

These datasets identify different fallacy types and range from 5 to 18 fallacies. \citet{alhindi-etal-2022-multitask} unified the fallacy types from the four schemes and introduced 28 fallacy types in one unified scheme. These dataset are different in size as they go from a few hundred examples (450-880) for \textsc{Climate, Covid} and \textsc{Argotario}, to a few thousands (4,500 to 5,100) for \textsc{Logic} and \textsc{Propaganda}. The total number of examples per fallacy type varies significantly as it ranges from less than 100 examples for some fallacies (e.g., \textit{False Analogy, Strawman, Whataboutism}) to more than 1,000 examples (e.g., \textit{Hasty Generalization, Name Calling or Labeling, Loaded Language}). Detailed numbers for each fallacy per dataset and split can be found in \citet{alhindi-etal-2022-multitask}. 

One main challenge in these datasets is the high imbalance frequency of classes in a high multi-class, and even multi-label task. The unified model presented by \citet{alhindi-etal-2022-multitask} improves the overall results but still performs much better on more frequent classes, thus we utilize data augmentation to address this challenge. In addition, two of the five fallacy datasets: \textsc{Propaganda} and \textsc{Climate} are from news articles where the fallacy is annotated at the sentence or fragment level. 

Therefore, we assess the benefit of providing additional context to the fallacious segment (sentence or fragment) by including the preceding or succeeding sentence when available.

\section{Synthetic Data Generation}
\label{sec:gen}
To generate additional examples for infrequent fallacy classes, we leverage {\normalsize \tt gpt-3.5-turbo} (henceforth called GPT-3.5), a conversational language model, as a data augmentation tool. We explore zero-shot, 1-shot, 2-shot, and 5-shot settings to generate examples that have not been seen in the original training data. These generated examples provide diversity and help address the data scarcity issue for less frequent fallacy types.

In order to understand the capabilities of pretrained large language models such as GPT-3.5 in producing synthetic data, we control the information provided in the prompts as follows: i) zero-shot prompts that have no fallacy example and ask the model to generate an example in one form (e.g., sentence, tweet, question-answer pair) for a certain fallacy provided in the prompt; ii) few-shot prompts that list the fallacy type and output form in addition to providing 1 to 5 examples for the given fallacy type in the prompt. The model is asked to generate the same number of examples given in the prompt (i.e. 1-shot prompt asks the model to generate 1 example, 5-shot ask for 5 new examples and so on); iii) few-shot-context prompts that provide the examples of fallacy and their wider context if available (previous and next sentence) and asks the language model to do the same by generating both examples for a certain fallacy and their contexts. Figure \ref{figure:1shot_ex} shows an example of the 1-shot-context prompt of the \textit{Irrelevant Authority} fallacy from the \textsc{Propaganda} dataset.

\begin{figure}[t!]
    \centering
    \small
    \begin{tabular}{ |p{7cm}| }

        \hline
        \\ \textbf{\color{teal} \underline{Prompt}} \\
            \\Your task is to perform the following actions:
            \\1. Read the sentence that has the fallacy of {\color{red} Irrelevant Authority} and its provided context. The sentence and context below are both delimited by <>. 
            \\2. Generate a similar example of one sentence and its context of the {\color{red} Irrelevant Authority} fallacy. The context should be a bigger chunk of text that includes the sentence,
            similar to the provided example sentence and context below.
            
            \\Your generated output should be in JSON format with the following keys:
            \color{blue} generated\_sentence, generated\_context.

            \\ {\color{olive} Sentence}: <Carlson cited Dr. Robert Epstein who has said, in Carlson’s words, ``Google alone could determine the outcome of almost any election just by altering its search selections and we would never know it.''>	
            \\{\color{olive} Context}: <Carlson cited Dr. Robert Epstein who has said, in Carlson’s words, “Google alone could determine the outcome of almost any election just by altering its search selections and we would never know it.''. Dr. Robert Epstein is a social scientist and an expert on Google. >

        \vspace{1mm}
        \\ \hline
        
        \\ \textbf{ \color{teal} \underline{Generated Output}} \vspace{1mm}
            \\ \{
            \\ {\color{blue} 'generated\_sentence'}: 'As a dog groomer, I can tell you that the earth is flat.',
            \\ {\color{blue} 'generated\_context'}: 'While most people accept that the earth is round, one dog groomer is convinced otherwise. As a dog groomer, I can tell you that the earth is flat. Dogs are great navigators, and they always sense when the ground is flat or sloping.'
            \vspace{1mm}
            \\ \}
            \vspace{1mm}
        \\ \hline
    \end{tabular}
    \caption{Example of 1-shot context prompt from the Propaganda dataset of a sentence that has the \emph{Irrelevant Authority} fallacy and its context.}
    \label{figure:1shot_ex}
\end{figure}

For all data augmentation settings, we generate the same number of examples per fallacy and thus study the quality of the synthetic data using different prompts for generation. The aim is to address data imbalance in these datasets so we exclude fallacies that are very frequent, especially if they only exist in one fallacy scheme (i.e. less diverse). Following this criteria, we exclude \textit{Loaded Language} and \textit{Name Calling or Labeling} that only appear in \textsc{Propaganda}. We also do not generate examples for \textit{Hasty Generalization} in a form similar to the \textsc{Logic} dataset, but we generate ones in Covid-19 and climate change domains since their respective datasets have this fallacy in very low counts.

For all generated fallacies we double the number of examples with respect to the number of original examples for a certain fallacy thus maintain comparable ratios of both original and synthetic data. Also, we noticed repetition in synthetic examples if we use a single original example multiple times in the prompts, which causes a drop in performance in the downstream task. Therefore, we cap the number of synthetic examples for each fallacy to 100 examples generated from each dataset. This changes the distribution of the training set by bringing the very infrequent classes closer to the overall average number of examples per class.

\section{Experimental Setup}
\label{sec:exp}
Similar to \citet{alhindi-etal-2022-multitask}, we use the T5 model \cite{raffel2020exploring}, a versatile text-to-text transformer, as the backbone for fallacy recognition by fine-tuning instruction-based prompts on all fallacy datasets. The prompts are designed to provide explicit instructions on identifying specific fallacies, enabling targeted learning within the model. This approach is inline with a large body of research that utilizes instruction-tuning of large language models on many tasks \cite{wei2021finetuned,sanh2022multitask}. We evaluate the proposed approach on the five fallacy datasets. We train the T5 model using a combination of the original labeled data and the generated examples from GPT-3.5. We compare the performance of the model under different settings, including zero-shot, 1-shot, 2-shot, and 5-shot scenarios, with and without additional context to understand the impact of prompt and data availability on fallacy recognition.

All fallacy examples, original and synthetic, are transformed into instruction-based prompts that are used to fine-tune the T5-3 Billion model (henceforth T53B) in a multitask fashion. The model and hyperparameters are fixed and we only change the training data that is fed into the model with the aim to study the ability of a smaller size model such as T53B to learn from manually annotated or crowdsourced data as well as synthetically generated data by a larger size model such as GPT-3.5. 

We show the results of all training conditions for the \textsc{Propaganda} and \textsc{Climate} datasets in Table \ref{tab:res1} and for the \textsc{Argotario}, \textsc{Logic}, and \textsc{Covid} datasets in Table \ref{tab:res2}. In both tables, we report the overall accuracy and macro F1 scores for each dataset as well as the F1 scores for each fallacy class. The results cover nine training conditions where we train on the original training set only (baseline-N (no-context)) similar to \cite{alhindi-etal-2022-multitask}, and baseline-C with context for datasets that are from news articles. This applies to the \textsc{Propaganda} and \textsc{Climate} where this context is available. The remaining seven training conditions all use a different form of data augmentation depending on the number of examples provided in the prompt during synthetic data generation which includes zero, one, two, or five examples. All data augmentation experiments are done with context (C columns) and no-context (N columns), except for zero-shot prompts that are done without context only. 

It was challenging for GPT-3.5 to provide usable examples with contexts in most cases without providing at least one example in the prompt for GPT-3.5 to follow. Therefore zero-shot prompts are only reported in no-context settings. We discuss in the next section the effect of both data augmentation and the addition of context in more details.

\section{Results}
\label{sec:res}
\begin{table*}[t!]
    \centering
    \begin{tabular}{|l|l||cc||c|cc|cc|cc|}
    \hline
         & & & &\multicolumn{7}{c|}{Data Augmentation}\\  
         Dataset&Fallacy &\multicolumn{2}{c||}{baseline} 
         &zero-shot &\multicolumn{2}{c|}{1-shot} &\multicolumn{2}{c|}{2-shot}
         &\multicolumn{2}{c|}{5-shot}\\ 
         & &N &C &N &N &C &N &C &N &C\\
    \hline \hline
        Propaganda 
        &Black and White Fallacy &.14 &.34  &\textbf{.39} &\textbf{.39} &.29  &.35 &.33  &.36 &.34\\
        &Causal Oversimplification &.34 &.27  &.41  &\textbf{.48} &.27  &.39 &.23  &.44 &.29\\
        &Doubt &.61 &.66  &.67  &.66 &.69  &.66 &\textbf{.71}  &.69 &.68\\
        &Exaggerate/Minimization &.34 &.32  &.44 &\textbf{.58} &.55  &\textbf{.58} &.47  &\textbf{.58} &.49\\
        &Fear or Prejudice &.49 &.44  &.49   &.54 &.49  &\textbf{.67} &.46  &.51 &.50\\
        &Flag-Waving &.64 &.67  &.67  &.68 &.67  &.67 &.67  &\textbf{.69} &\textbf{.69}\\
        &Irrelevant Authority &.26 &.30  &.44  &.46 &.44  &\textbf{.47} &.40  &.41 &.38\\
        &Loaded Language &.79 &.76  &.81  &\textbf{.83} &.81  &\textbf{.83} &.79  &\textbf{.83} &.80\\
        &Name Calling, Labeling &.83 &.79  &.83   &.85 &.82  &.85 &.81  &\textbf{.86} &.81\\
        &Red Herring &0 &0  &0   &\textbf{.29} &.22  &0 &0  &0 &0\\
        &Reductio Ad Hitlerum &.17 &.18  &.29  &.40 &.27  &\textbf{.44} &.25  &.40 &.22\\
        &Slogans &.49 &.45  &.59  &.56 &.52  &.55 &.51  &\textbf{.67} &.48\\
        &Strawman &0 &0  &0   &0 &0  &0 &0  &0 &0\\
        &Thought-Termin. Cliches &.29 &.34  &.29   &\textbf{.50} &.40  &.36 &.41  &.38 &.39\\
        &Whataboutism &0 &0  &.29  &\textbf{.63} &.62  &.53 &.53  &.48 &.47\\
     \hline
        &Accuracy &.68 &.67  &.71  &\textbf{.74} &.71  &.73 &.70  &\textbf{.74} &.70\\
        &Macro &.36 &.37  &.44  &\textbf{.52} &.47  &.48 &.44  &.49 &.44\\
    \hline \hline
        Climate&Causal Oversimplification &.35 &.33  &.40 &.53 &.32  &.42 &.30  &\textbf{.60} &.37\\
        &Cherry Picking &.44 &.41  &.43  &\textbf{.48} &.44  &.43 &.41  &.46 &.45\\
        &Evading Burden of Proof &0 &0  &0  &\textbf{.17} &.12  &0 &.10  &0 &0\\
        &False Analogy &0 &0  &.36  &\textbf{.62} &.18  &.35 &.17  &.43 &.17\\
        &Hasty Generalization &0 &0  &0   &0 &0  &0 &0  &0 &0\\
        &Irrelevant Authority &.22 &.25  &.31   &.31 &.31  &.31 &.31  &\textbf{.43} &.33\\
        &Red Herring &0 &0  &.12   &.11 &0  &\textbf{.18} &0  &\textbf{.18} &0\\
        &Strawman &.22 &0  &.40   &.40 &.40  &.36 &.50  &\textbf{.55} &.40\\
        &Vagueness &.37 &.39  &.34 &\textbf{.40} &.29  &.36 &.36  &.36 &.24\\
    \hline
        &Accuracy &.30 &.28  &.34 &\textbf{.40} &.29  &.34 &.29  &.39 &.30\\
        &Macro &.18 &.15  &.26  &\textbf{.33} &.23  &.27 &.24  &\textbf{.33} &.22\\
    \hline 
    \end{tabular}
    \caption{F1 scores on the Propaganda and Climate datasets using multitask training of T53B model. \textbf{N}: no context to the fallacious segment added. \textbf{C}: context of previous and next sentence to the fallacious segment provided.}
    \label{tab:res1}
\end{table*}

\begin{table*}[t!]
    \centering
    \begin{tabular}{|l|l||cc||c|cc|cc|cc|}
    \hline
         & & & &\multicolumn{7}{c|}{Data Augmentation}\\ 
         Dataset&Fallacy &\multicolumn{2}{c||}{baseline} 
         &zero-shot &\multicolumn{2}{c|}{1-shot} &\multicolumn{2}{c|}{2-shot}
         &\multicolumn{2}{c|}{5-shot}\\ 
         & &N &C &N &N &C &N &C &N &C\\
    \hline \hline
        Argotario &Ad Hominem &.59 &.63  &.63 &\textbf{.71} &.64  &.62 &.64  &.65 &.63\\
        &Emotional Language &.64 &.68  &.70  &\textbf{.71} &.70  &.67 &.69  &.60 &.65\\
        &Hasty Generalization &.46 &.44  &.51 &.47 &.54  &.47 &.52  &\textbf{.55} &.49\\
        &Irrelevant Authority &.71 &.72  &\textbf{.80} &.75 &.78  &.74 &.77  &.75 &.78\\
        &Red Herring &.32 &.42  &.47  &.50 &.46  &.44 &.51  &\textbf{.53} &.52\\
    \hline
    &Accuracy &.56 &.57  &.61  &.61 &.60  &.59 &\textbf{.63}  &.61 &.62\\
    &Macro &.54 &.58  &.62  &.61 &.61  &.58 &\textbf{.63}  &.61 &.61\\
    \hline \hline
        Logic&Ad Hominem &.77 &.81 &.87 &\textbf{.88} &\textbf{.88}  &\textbf{.88} &.85  &.86 &\textbf{.88}\\
        &Ad Populum &.81 &.80  &.82  &\textbf{.89} &.87  &.86 &.86  &\textbf{.89} &.85\\
        &Black and White Fallacy &.84 &.84  &.91  &.91 &.89  &\textbf{.92} &.89  &.91 &\textbf{.92}\\
        &Causal Oversimplification &.65 &.70  &.81 &.79 &\textbf{.82}  &.81 &.80  &.78 &.79\\
        &Circular Reasoning &.57 &.56  &.68 &\textbf{.84} &\textbf{.84}  &.80 &.77  &.76 &.83\\
        &Deductive Fallacy &.32 &.29  &.48  &\textbf{.69} &.57  &.57 &.54  &.56 &.57\\
        &Emotional Language &.55 &.53  &.65   &.76 &\textbf{.77}  &.72 &.74  &.71 &.68\\
        &Equivocation &.22 &0  &.27  &\textbf{.57} &.43  &.55 &.39  &.43 &.42\\
        &Fallacy of Extension &.08 &.04  &.48   &\textbf{.68} &\textbf{.68}  &.64 &.60  &.58 &.64\\
        &Hasty Generalization &.64 &.63  &.72   &\textbf{.80} &.75  &.77 &.75  &.77 &.79\\
        &Intentional Fallacy &.09 &.15  &.16   &\textbf{.55} &.48  &.46 &.33  &.35 &.33\\
        &Irrelevant Authority &.56 &.54  &.61   &\textbf{.74} &.68  &.68 &.72  &.68 &.66\\
        &Red Herring &.24 &.30  &.58   &\textbf{.78} &.67  &.67 &.61  &.65 &.62\\
     \hline
        &Accuracy &.58 &.58  &.68   &\textbf{.79} &.76  &.75 &.73  &.73 &.74\\
        &Macro &.45 &.48  &.62   &\textbf{.76} &.72  &.72 &.68  &.69 &.69\\
    \hline \hline
        Covid&Causal Oversimplification &.45 &.53  &.40   &.56 &\textbf{.59}  &.53 &.53  &.50 &.50\\
        &Cherry Picking &.35 &.37  &.37   &.31 &.36  &.28 &.34  &\textbf{.38} &\textbf{.38}\\
        &Evading Burden of Proof &0 &0  &.31   &.45 &.53  &.46 &\textbf{.57}  &.49 &.40\\
        &False Analogy &.33 &\textbf{.50}  &.25   &.29 &.29  &.25 &.29  &.29 &.25\\
        &Hasty Generalization &.17 &0  &.11   &.17 &.16  &.11 &\textbf{.25}  &.10 &.11\\
        &Irrelevant Authority &0 &0  &0   &0 &0  &0 &0  &0 &0\\
        &Red Herring &0 &0  &0  &0 &0  &0 &0  &0 &0\\
        &Strawman &0 &0  &0  &\textbf{.20} &0  &0 &.17  &.17 &.15\\
        &Vagueness &.09 &0  &.09   &.27 &\textbf{.33}  &.22 &.30  &.15 &.19\\
    \hline
        &Accuracy &.23 &.25  &.26   &.30 &.34  &.27 &\textbf{.36}  &.31 &.30\\
        &Macro &.16 &.15  &.17   &.25 &.25  &.20 &\textbf{.27}  &.23 &.22\\
    \hline
    \end{tabular}
    \caption{F1 scores on the Argotario, Logic and Covid datasets. \textbf{N}: no additional context provided. \textbf{C}: context of previous and next sentence provided where available  (Propaganda and Climate only and added with no-context training sets of the three datasets shown).}
    \label{tab:res2}
\end{table*}

\paragraph{Data Augmentation.}
Adding synthetic data to the original data improves the results over the baselines where only the original training data is used regardless of the data augmentation method. This is true for both the overall accuracy and macro-F1 scores in all five datasets as shown in Tables \ref{tab:res1} and \ref{tab:res2} whether the context is provided or not. Interestingly, 1-shot prompts seem to yield the best results when compared to both zero-shot and other few-shot settings. This results is counter to what we initially expected. We hypothesized that 5-shot prompts that have five examples of a fallacy and ask GPT-3.5 to generate five similar examples to yield synthetic data that is more generic to the fallacy (the one factor that is common among the five examples in the prompt), and therefore would help train a model for fallacy recognition to be more resilient. However, it seems that 5-shot prompts generate synthetic examples that are more similar to the original data (more in Section \ref{sec:analysis}), which consequently makes the synthetic data generated by 1-shot prompts better for the downstream task.



\paragraph{Per-Class Results.} 
Some fallacies show larger gains after data augmentation compared to others. This is true in the \textsc{Logic} dataset where the \textit{Equivocation} and \textit{Fallacy of Extension} are among fallacies with the biggest gains over baselines. These two fallacies are also the least frequent in the \textsc{Logic} dataset and thus the impact of data augmentation is bigger. The diversion fallacies in \textsc{Propaganda} e.g., \textit{Red Herring, Strawman, Whataboutism} are particularly challenging in baseline settings due to their low counts and complexity since they could require external information to the fallacious segment to be properly recognized, which is especially the case for \textit{Strawman} where all models fail to make any correct prediction with or without data augmentation. However, for \textit{Red Herring} and \textit{Whataboutism}, significant gains are observed with data augmentation particularly for \textit{Whataboutism} in 1-shot settings where the f1-scores jumps to 0.63 compared to 0 in the baseline models. 

Some fallacy types are challenging to detect in some datasets but not as challenging in other datasets. This is mainly due to their format in a particular dataset, frequency, and the fallacies they are listed with in the prompt at inference time. For example, \textit{Red Herring} is easier to detect in \textsc{Argotario} and \textsc{Logic} even by the baseline model due to a lower number of fallacies in \textsc{Argotario}, and the lack of other diversion fallacies in these two scheme, which makes \textit{Red Herring} more distinct than the other fallacies and easier to distinguish. However, for \textsc{Propaganda} and \textsc{Climate}, the baselines get 0 f1-scores for \textit{Red Herring} and data augmentation helps in improving the results to 0.29 in 1-shot for \textsc{Propaganda} and 0.18 in 2-shot and 5-shot settings in the \textsc{Climate} dataset. 

Some fallacy types remain challenging to detect with any kind of data augmentation, such as \textit{Strawman} in \textsc{Propaganda}, and \textit{Hasty Generalization} in \textsc{Climate} given their low counts in the test set (e.g., 2-5 examples) and therefore the test sets might have one particular form of this fallacy rather than represent the fallacy type in general. Having a train-test split that can truly evaluate the performance of machine learning models for this task is not trivial due to the high number of classes, the severe data imbalance, the quality of labels, and the inherent subjectivity and overlap between some fallacies. Revisiting the splits would hinder the ability of direct comparison with previous work on these datasets and therefore it is not included in this paper but worth re-examination in future work.



\begin{figure*}[th]
\begin{subfigure}{\textwidth}
  \includegraphics[width=0.92\linewidth]{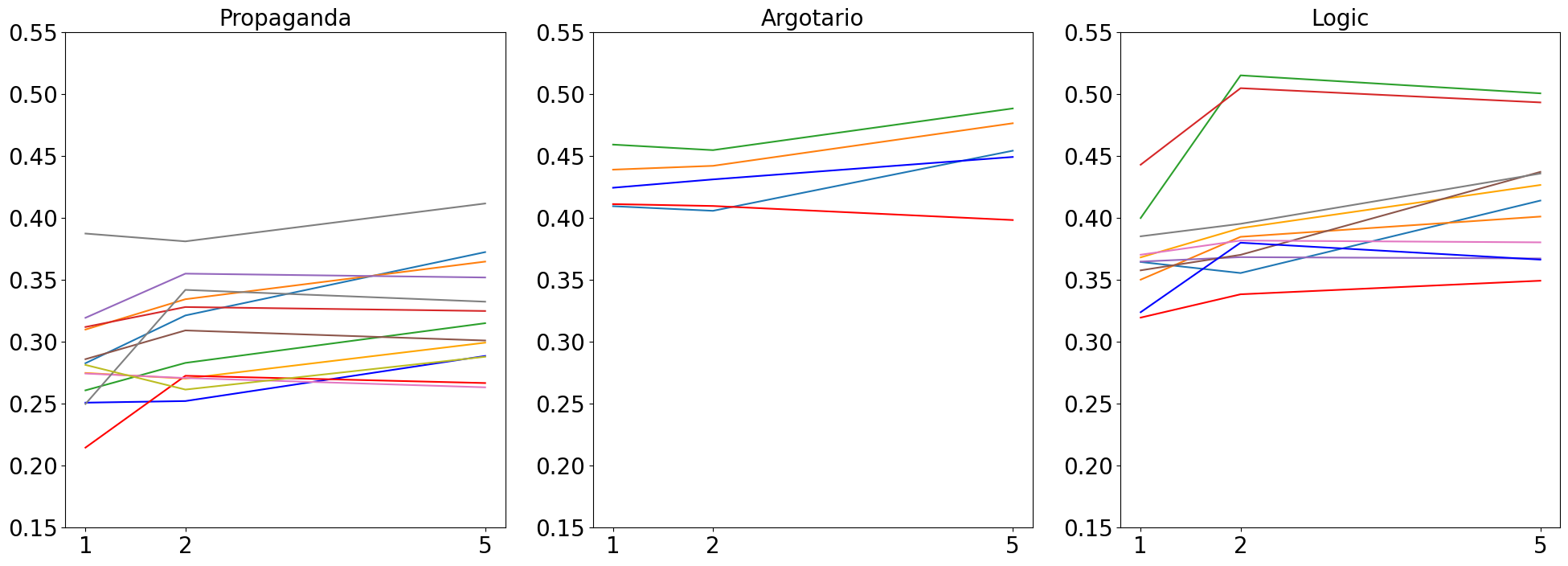}
  \label{fig:analysis1}
\end{subfigure}
\\
\begin{subfigure}{\textwidth}
  \centering
  \includegraphics[width=\linewidth]{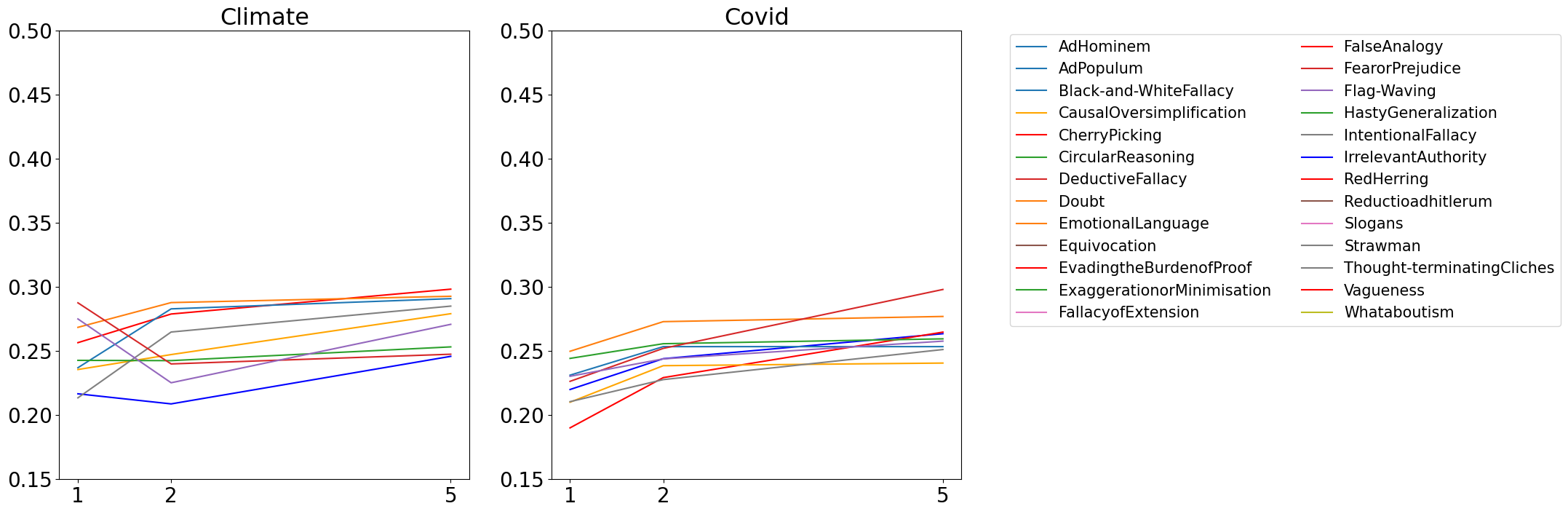}
  \label{fig:analysis2}
\end{subfigure}
\caption{Average BLEURT score (y-axis) between original and synthetic data for each fallacy type in few-shot prompts (x-axis: 1-shot, 2-shot, 5-shot).}
\label{fig:analysis}
\end{figure*}

\paragraph{Effect of Additional Context.}
The use of context during training is different for \textsc{Propaganda} and \textsc{Climate} in Table \ref{tab:res1} compared to the other three datasets shown in Table \ref{tab:res2}. The difference between each N and C columns in Table \ref{tab:res1} is rather than only providing a fallacious segment, we provide a wider context window of the previous and next sentence when available for the two datasets listed in the table. However, there is no difference in the training data between the N and C columns for the three datasets listed in Table \ref{tab:res2} i.e. \textsc{Argotario}, \textsc{Logic}, and \textsc{Covid}. The only difference is that they are combined in the multitask training model with two other datasets (\textsc{Propaganda} and \textsc{Climate}) where the context is provided. Since the training is done on all datasets combined with some overlap between fallacy types across datasets, we report the results on the \textsc{Argotario}, \textsc{Logic}, and \textsc{Covid} datasets for context-based experiments on all five datasets.

With minor exceptions (e.g., \textit{Doubt} in \textsc{Propaganda}, \textit{Vagueness} and overall scores in \textsc{Covid}), adding context does not improve the results for fallacy recognition. This could be related to the fact that some fallacy types require different context than others. For example, \textit{Cherry Picking} requires understanding of the trend and \textit{Strawman} requires the retrieval of the original argument, while \textit{Evading the Burden of Proof} needs information regarding the structure of the argument to assess its validity \cite{goffredo2022fallacious,alhindi-etal-2021-fact}. Therefore, a unified form of context across twenty-eight fallacy types does not have consistent improvement over experiments conducted under similar conditions.


\paragraph{Overall Observations.} There are two observations that are consistent across all results. First, data augmentation through large language models helps train smaller models on more data that is beneficial to fallacy recognition. Second, simple context of previous or next sentence does not provide valuable insight for this task. Thus, the customization of the type of context based on the fallacy type is needed. 

\section{Original and Synthetic Data Similarity}
\label{sec:analysis}
In order to understand the reason for 1-shot prompts to generate synthetic data that is more beneficial to the task, we analyze the similarity between the generated data and the original training examples shown at the prompts. 
For similarity, we use BLEURT score as it has the most consistent results with human evaluation \cite{sellam-etal-2020-bleurt}. We calculate BLEURT score for each original-synthetic example pair where the original example is the one used in the prompt to generate the synthetic example. Thus, we only report similarity in the 1-shot, 2-shot, and 5-shot prompts. For the 2-shot and 5-shot, we report the maximum score for a generated example with respect to all original examples included in the prompt. 

Figure \ref{fig:analysis} shows average BLEURT scores for each fallacy in all five datasets. We notice high similarity scores in \textsc{Argotartio} and \textsc{Logic} that range between 0.50 and 0.30 and much lower scores for \textsc{Climate} and \textsc{Covid} that range between 0.18 and 0.30. This shows that it is harder to produce synthetic examples that are similar to naturally-occurring examples in news and social media. 

Common across all datasets is that on average 1-shot prompts for most fallacies tend to produce less similar examples to the ones included in the prompts when compared with 2-shot and 5-shot prompts. However, the synthetic data from the 1-shot prompts is more useful to training a model for fallacy recognition that is tested on human-annotated data. After closer examination of synthetic data from 2-shot and 5-shot, while they are more similar to the original data on average, they tend be more similar to one of the 2 example (or 5 examples) provided in the prompt and thus suffer from repetition of one form, which makes the overall synthetic data less diverse. This shows the sensitivity of the LLM to the information in the prompt and possibly to the order of the provided few-shot examples. Therefore, asking an LLM to generate multiple synthetic examples in one instruction could lead it to generate ones that are more similar to each other even if the original data in the prompt include multiple human-annotated examples.




\section{Related Work}
\label{sec:rw}
We provide an overview of the literature that discusses the use of LLMs for data augmentation, and the literature on the development of models and resources for fallacy recognition.

\subsection{Data Augmentation} With the significant focus on the development of generative large language models (LLMs) in recent years \cite{brown2020language, zhang2022opt, chowdhery2023palm, touvron2023llama, achiam2023gpt,jiang2023mistral}, there has been an increase in the utilization of these models to annotate data \cite{feng-etal-2021-language, chen-etal-2023-places, he2023annollm, bansal2023large, zhang2023llmaaa}, or generate additional data instances that can be added to existing training sets for various tasks \cite{kumar-etal-2020-data, schick-schutze-2021-just, wang2023gpt, wang2021towards, ye-etal-2022-zerogen, gao2022self, sahu2023promptmix}. LLMs are particularly useful for data augmentation in specialized domains or complex tasks where human labels are difficult or expensive to collect at scale \cite{ding2024data}. Previous work include using LLMs for annotation following an explain-then-annotate approach \cite{he2023annollm}, a framework for LLMs as active annotators \cite{zhang2023llmaaa}, and a sampling strategy to find the most informative inputs to annotate by LLMs \cite{bansal2023large}. \citet{moller2023prompt} instructs LLMs to generate examples similar to a provided example from one class and uses those for few-shot classification. \citet{sahu2023promptmix} generates challenging augmentations near class boundaries and instructs LLMs to relabel these augmentations. 

For dialogue, \citet{feng-etal-2021-language} uses GPT for dialogue summarization, \citet{chen-etal-2023-places} uses expert written conversation to synthesize social conversation, \citet{chintagunta2021medically} uses GPT-3 to create medical summaries in dialog setting then trains models on a mix of synthesized and gold-labeled data which scales the results to 30x gold-labeled examples, and AugESC \citet{zheng2023augesc} finetunes an LM and prompts it to complete dialogues for emotional support conversations. We adopt similar strategies for fallacy recognition that presents a unique set of challenges.

\subsection{Fallacy Recognition}
In addition to the five fallacy datasets mentioned in Section \ref{sec:dataset} that cover fallacy in dialogue \cite{habernal-etal-2017-argotario}, propaganda \cite{da-san-martino-etal-2019-fine}, educational websites \cite{jin-etal-2022-logical} and (mis/dis)-information \cite{musi2022developing,alhindi-etal-2022-multitask}, there is other work that extends these datasets in the same domain or cover other domains and genres. \citet{habernal-etal-2018-name} created a dataset for the ad hominem fallacy from the "Change My View" subreddit, while \citet{sahai-etal-2021-breaking} used Reddit to create a dataset with eight fallacy types. In addition, fallacy in dialogue include datasets in political debate \cite{goffredo2022fallacious,goffredo2023argument}, and the evaluation of the robustness of LLMs against logical fallacies \cite{payandeh2023susceptible}. 

Work on propaganda include covering additional propaganda techniques that go up to a total of 23 techniques \cite{piskorski-etal-2023-multilingual}, the detection of propaganda techniques in code-switched text \cite{salman2023detecting}, using discourse to detect propaganda \cite{chernyavskiy2024unleashing}, and the study of framing and how it relates to persuasion and propaganda \cite{sajwani2024frappe}. 

A shared task at the CheckThat! Lab at CLEF 2024 extends the coverage of propaganda techniques to 13 languages\footnote{\href{https://checkthat.gitlab.io/clef2024/task3/}{checkthat.gitlab.io/clef2024/task3/}}, while other shared tasks cover text and memes in multilingual settings\footnote{{\href{https://propaganda.math.unipd.it/semeval2024task4/}{propaganda.math.unipd.it/semeval2024task4/}}}, and in Arabic\footnote{\href{https://araieval.gitlab.io/}{araieval.gitlab.io/}} \cite{hasanain2023araieval}.


We build on the line of work that uses language models for generating additional training data. Our work differs form previous work in the following aspects: i) we particularly focus on the ability of using synthetic data generated by language models to address data imbalance challenges, ii) we use zero-shot and few-shot settings to generate synthetic data but use full-shot training on a mix of original and synthetic data for the downstream task, and iii) we tackle a challenging task of fallacy recognition to understand the gains from using large language models for data augmentation.

\section{Conclusion and Future Work}
\label{sec:conc}
Fallacy recognition remains a challenging problem due to the high number of classes, severe data imbalance and the need in some cases for external information to the fallacious segment. To mitigate the effect of data imbalance, we studied the capabilities of large language models to generate synthetic data that can be used to train smaller models on a combination of original and synthetic data for fallacy recognition across multiple tasks. The main observation is that data augmentation through large language models is beneficial for this task. 

However, the conditions under which the data is generated impacts the quality of the synthetic data significantly. Providing one example in the prompt (1-shot) for a certain fallacy from the original data and asking GPT-3.5 to generate a similar example results in synthetic data that is more beneficial to downstream models for fallacy recognition tested on data from the same distribution. The value in having synthetic data that is less similar to the original training data and possibly more generic to the task needs to be tested on data from unseen fallacy schemes or domains, which presents a potential avenue for future work. Overall, large language models show great potential to generate additional training data for the task of fallacy recognition, which can be used to train smaller size open-source models for this task. 

In future work, we want to test the resilience of data augmentation on out-of-domain test sets such as fallacy in political debates. Also, we want to study the ability of LLMs to generate examples that could be labeled by multiple fallacies and train machine learning for this tasks with a multi-labeling objective. Finally, we want to experiment with the ability of LLMs to provide more useful context for fallacy recognition.

\section*{Limitations}
This work addresses challenges related to datasets with imbalance class ratios in high multi-class classifications using data augmentation generated by large language models. However, this work does not address other challenges in fallacy recognition. 

These include the incorporation of external knowledge to the fallacious segment which is essential in detecting some of the diversion fallacies such as {\it Cherry Picking} that requires an understanding of the general trend for the issue at hand, or {\it Strawman} that requires the retrieval of the original argument.

In addition, this work assumes a single fallacy label for each segment of text. However, in reality fallacies can overlap and thus handling the multi-label aspect of this task is not covered in this work. 

Also, this work focuses on the classification of fallacy types and not the detection of fallacy and separating it from non-fallacious text which is a challenging task \cite{goffredo2023argument}. 

Finally, labeling fallacy by humans is inherently subjective and thus concurrent work suggests incorporating subjectivity in fallacy labels \cite{helwe2023mafalda}, and thus treating human annotations as certain gold labels might provide a limited perspective for fallacy recognition models.

\section*{Ethics and Broader Impact}
Using large language models to generate examples of various fallacy types comes at a risk of producing impolite, foul or hateful language. We have inspected a sample of the synthetic data and modified the prompts to minimize these aspects in the generated data. However, it is hard to guarantee the nonexistence of harsh language in data from large language models at scale. Some fallacy techniques in the datasets used in this paper have harsh or impolite language by definition e.g., {\it Name Calling, Labeling} and thus cannot be avoided. Also, studying fallacy and training machine learning models for fallacy recognition could potentially lead to the promotion of the topic and the misuse of these models in various ways. 

While we acknowledge the risks, we believe this study contributes to increasing the awareness of fallacious techniques for both readers and writers and can better equip them with proper tools to increase their immunity against potential harms with the overall objective of increasing digital literacy for both content-creators and content-consumers. 


\bibliography{anthology,custom-fixed}
\bibliographystyle{acl_natbib}

\appendix



\end{document}